\documentclass[conference]{IEEEtran}
\IEEEoverridecommandlockouts
\usepackage{bm}
\usepackage{fancyvrb,cprotect}
\usepackage{mathtools}
\usepackage{algorithmic}
\usepackage{graphicx}

\usepackage{amsmath}
\usepackage{amsthm}
\usepackage{amssymb}
\usepackage{adjustbox}
\usepackage[caption=false]{subfig}
\usepackage[utf8x]{inputenc}
\usepackage{multirow,booktabs}

\def\BibTeX{{\rm B\kern-.05em{\sc i\kern-.025em b}\kern-.08em
    T\kern-.1667em\lower.7ex\hbox{E}\kern-.125emX}}
\begin{document}

\title{VS-Net: Multiscale Spatiotemporal Features for Lightweight Video Salient Document Detection
\thanks{Supported by Ministry of Electronics and Information Technology (MeiTy), Government of India and IIT Bhilai Innovation and Technology Foundation (IBITF) under the project entitled "Blockchain and Machine Learning Powered Unified Video KYC Framework"}
}

\author{\IEEEauthorblockN{1\textsuperscript{st} Hemraj Singh}
\IEEEauthorblockA{\textit{Dept. of Computer Sc. \& Engg.} \\
\textit{National Institute of Technology}\\
Warangal, Telangana, India \\
hs720079@student.nitw.ac.in}
\and
\IEEEauthorblockN{2\textsuperscript{nd} Mridula Verma}
\IEEEauthorblockA{\textit{Institute for Development and Research} \\
\textit{ in Banking Technology}\\
Hyderabad, Telangana, India \\
vmridula@idrbt.ac.in}
\and
\IEEEauthorblockN{3\textsuperscript{rd} Ramalingaswamy Cheruku}
\IEEEauthorblockA{\textit{Dept. of Computer Sc. \& Engg.} \\
\textit{National Institute of Technology}\\
Warangal, Telangana, India \\
rmlswamy@nitw.ac.in}
}

\maketitle

\begin{abstract}
Video Salient Document Detection (VSDD) is an essential task of practical computer vision, which aims to highlight visually salient document regions in video frames. Previous techniques for VSDD focus on learning features without considering the cooperation among and across the appearance and motion cues and thus fail to perform in practical scenarios. Moreover, most of the previous techniques demand high computational resources, which limits the usage of such systems in resource-constrained settings. To handle these issues, we propose VS-Net, which captures multi-scale spatiotemporal information with the help of dilated depth-wise separable convolution and Approximation Rank Pooling. VS-Net extracts the key features locally from each frame across embedding sub-spaces and forwards the features between adjacent and parallel nodes, enhancing model performance globally. Our model generates saliency maps considering both the background and foreground simultaneously, making it perform better in challenging scenarios. The immense experiments regulated on the benchmark MIDV-500 dataset show that the VS-Net model outperforms state-of-the-art approaches in both time and robustness measures.
\end{abstract}


\section{Introduction} 
\begin{figure}[ht]
  \centering
  \subfloat[Complex Scene]{\label{figur:1}\includegraphics[width=25.5mm,height=14mm]{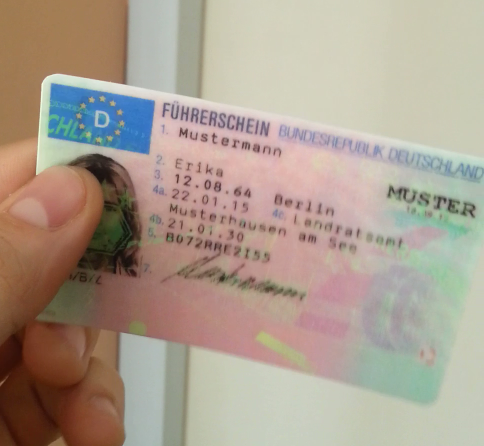}}\
  \subfloat[Keyboard Scene]{\label{figur:2}\includegraphics[width=25.5mm,height=14mm]{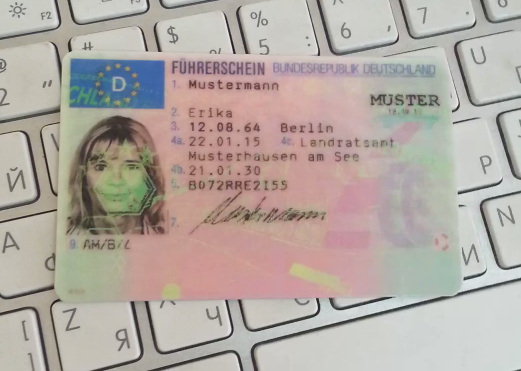}}\
  \subfloat[Partial Scene]{\label{figur:4}\includegraphics[width=25.5mm,height=14mm]{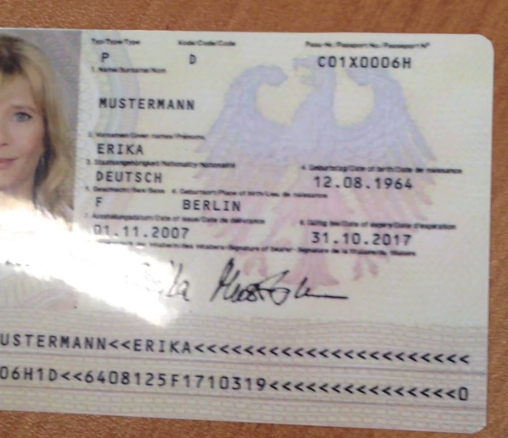}}\\
  \subfloat[Noise]{\label{figur:1}\includegraphics[width=25.5mm,height=14mm]{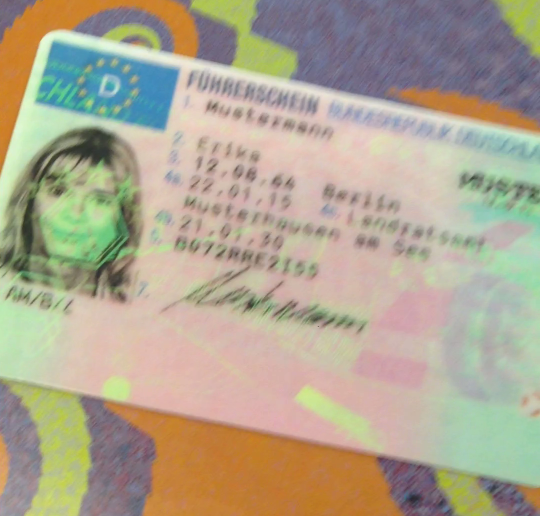}}\
  \subfloat[Motion Blur]{\label{figur:2}\includegraphics[width=25.5mm,height=14mm]{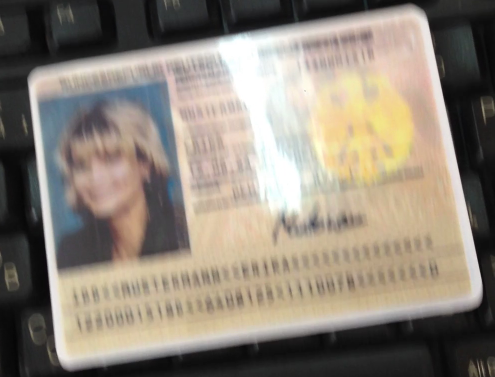}}\
  \subfloat[Illumination]{\label{figur:4}\includegraphics[width=25.5mm,height=14mm]{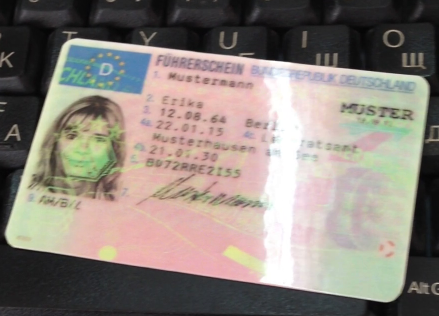}}
  \caption{Challenging scenarios from  MIDV-500 \cite{arlazarov2019midv} dataset.}
  \label{figure:one}
\end{figure}

Video Salient Document Detection (VSDD) is an essential task in several real-world applications, such as video document recognition \cite{burie2021deep}, video document compression \cite{sheshku2020houghencoder}, video document captioning \cite{schreiber2017deepdesrt} and many more. In real-life scenarios, a number of challenges appear due to an unconstrained environment (as shown in \textit{Fig.}\ref{figure:one}). Current state-of-the-art (SOTA) models \cite{sun2020content,burie2021deep} utilize non-selective attentional resources in the dynamic scenes. They employ limited static features and thus face difficulties in detecting the intended object in multiple real-world scenarios. 

Most of the existing VSDD models \cite{liu2021novel, kompella2021semi, huang2021novel} extract the spatial features separately using a computationally costly process and then integrate them to generate a spatial saliency map. Later, they use a different method to extract the refined spatial-temporal features. Segregating these two steps reduces the quality of the generated frame. This segregation also fails to capture the longer-term motion arrangement, which links with some actions. 

One of the most popular model for video analytics is U-Net \cite{liu2019u}, which extends the temporal dimension for substituting 2D filters with 3D filters and produces a little benediction of annotated videos to help the 3D convolution layers. Sebastian et al. \cite{schreiber2017deepdesrt} proposed DeepDeSRT, an end-to-end system for table understanding in document images and detecting PDF documents. But it failed in the video datasets due to poor handling of temporal features. To solve these problems, Recurrent Neural Networks (RNNs) \cite{kompella2021semi} used memory cells for the long-term pattern, which parses the video frames sequentially and encodes the information. The LSTMs use convolutional neural networks \cite{rhanoui2019cnn} and output in the form of action labels or video specifications. The Autoencoder LSTM model \cite{sun2020content} is proposed to use either an instant or the next frame for accurate reconstruction. Tenet model in \cite{ren2020tenet} acquired salient object detection metrics and performed unsupervised training on CNN. Sheshkus et al. \cite{sheshku2020houghencoder} proposed HoughEncoder neural network architecture and performed Fast Hough Transform to calculate low-level features for the image semantic segmentation task, however, failed to resolve challenges due to an unconstrained environment. In \cite{le2018video} a spatiotemporal conditional random field  is proposed to establish the relationships between local and global context regions, but the method failed to extract the high-level features. Wujie Zhou et al. \cite{zhou2020salient} designed a convolution residual module to send equally distributed feature maps between the encoder and the decoder but failed due to long-range skip connections. The recent method \cite{huang2022inversion} proposed Vnet to optimize the skip connection but failed to combine multilevel feature information.

We propose a novel VS-Net model to overcome these problems, which utilizes the separable convolution in the combination of the variational encoder to extract the key features from each frame across embedding sub-spaces and forward the features between adjacent and parallel nodes. Our model extracts the spatiotemporal features locally and makes better predictions globally. The main contributions of our paper are given below: 
\begin{itemize}[noitemsep, nolistsep]
\item We design a novel spatiotemporal-based VS-Net model with separable convolutions in variational autoencoder architecture (VAE) \cite{li2019supervae}, which reduces the skip-connection between two nodes and generates the generalized latent space vector.
\item We utilize the Approximation Rank Pooling (ARP) \cite{bilen2017action}, which takes input features from separable convolutions intermediate layers to train the VS-Net model. It provides low-rank approximation features to preserve their temporal locality. 
\item We have conducted experiments with MIDV-500 \cite{arlazarov2019midv} dataset and demonstrated that VS-Net performs better in terms of both efficiency and time.
\end{itemize} 

\section{Proposed Methodology} 
\subsection{VS-Net Architecture}
\begin{figure*}[ht]
\centering
\includegraphics[width=10cm]{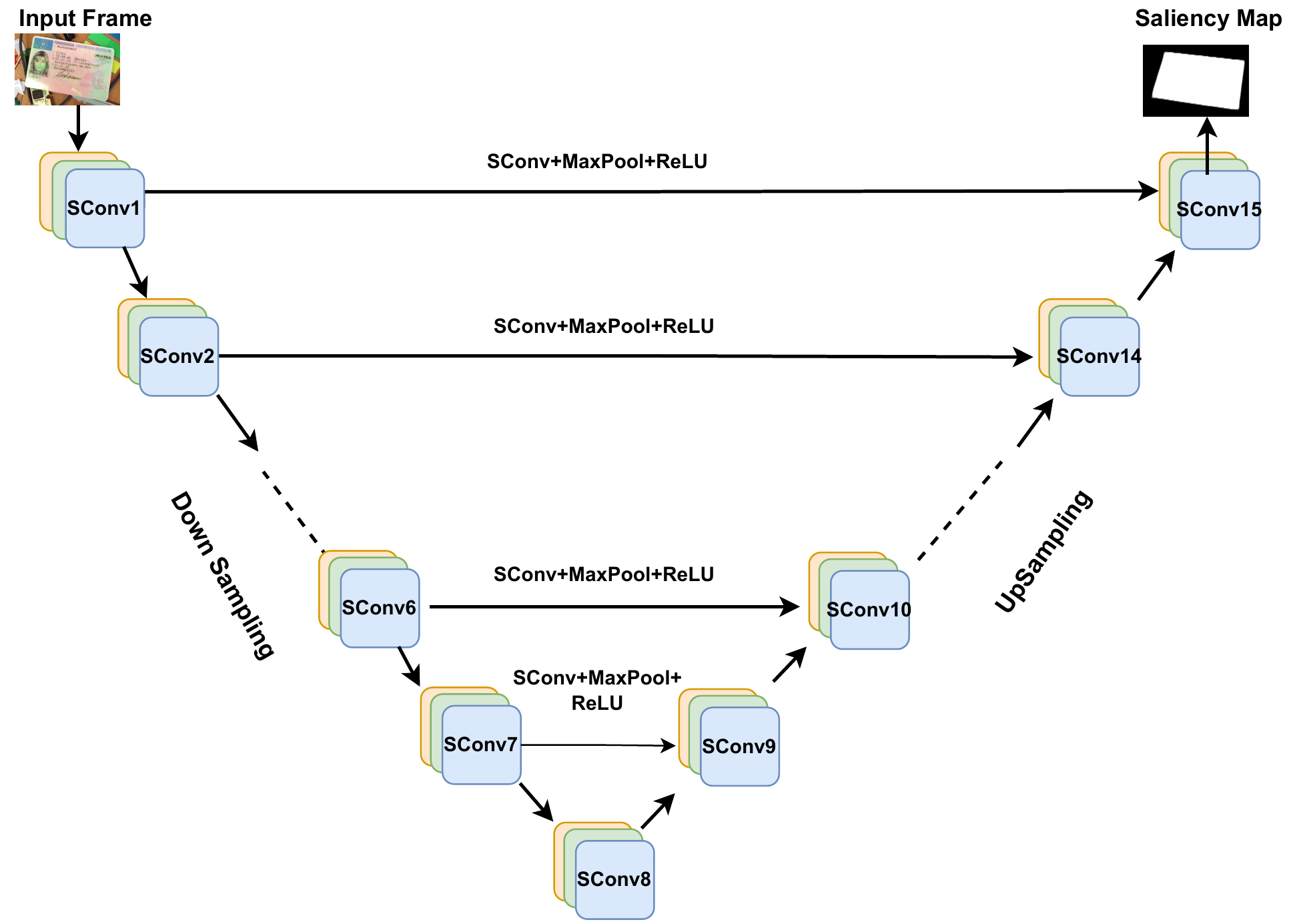}
\caption{Architecture of the proposed VS-Net.}
\label{fig:two}
\end{figure*}

Based on prior knowledge, spatial and temporal-based methods can capture better location information and preserve location boundaries than pixel-wise CNN methods. Therefore, we design a novel spatial and temporal-based VS-Net model with separable convolutions \cite{liu2021samnet} in variational auto-encoder architecture (VAE) \cite{li2019supervae}, which reduces the skip-connection between two nodes and generates the generalized latent space vector (shown in fig.\ref{fig:two}). 

Given a sequence of input frames ($S_{n} | n = 1, 2, 3, \cdot\cdot\cdot, N$), and corresponding ground-truth maps ($G_{n} | n = 1, 2, 3, \cdot \cdot \cdot , N$) are first passed into the VS-Net model to extract the spatial and temporal features, which uses pretrained weights of ResNet50 \cite{kompella2021semi}. Our proposed model has two branches with different purposes. The first is the down-sampling operation performing top to bottom, extracting the spatial and temporal features from each node and reducing the feature vectors' dimension. The second is upsampling operation from bottom to top, which decodes the spatial-temporal latent space and enhances the feature vectors. At last, we combine feature vectors from previous nodes and parallel nodes.

During the down-sampling, we perform separable convolution operation with $3 \times 3$ filters on input frames to extract spatial and temporal features, which have rich spatial and temporal information. Then max-pooling operation with $2 \times 2$ filters succeeded by a ReLU activation operation performs to downgrade the dimension of the features vectors and generate the latent spatial and temporal map. 
\begin{equation}
    S_{d} \sim Down(S_{n})= SConv(S_{1}, S_{2} ,S_{3} ,S_{4} ,\cdot\cdot\cdot, S_{n}),
\end{equation}
where \textit{SConv} is a separable convolution operation.

Before the up-sampling, we perform a convolution operation using $1 \times 1$ filters with learnable weight $\theta$ and applied the ReLU activation function to reduce the dimension of the latent features vectors and apply a dropout operation to dropout the 50$\%$ neurons for generating the latent space of spatial and temporal features vectors.
   \begin{equation}
   S_{l} = Dropout(ReLU(Conv(S_{d}, \theta))).
  \end{equation}
During the up-sampling operation from bottom to top, we decode the latent space vectors using separable convolution layers with $3 \times 3$ filters, succeeded by up-sampling layers and ReLU activation operation to reconstruct the dimension of the features vectors.
\begin{equation}
    S_{up}\sim UP(S_{l})= SConv(S_{1},S_{2},S_{3},S_{4},\cdot\cdot\cdot, S_{n}),
\end{equation}
where \textit{SConv} is a separable convolution layers with $3 \times 3$ filters. We extract the spatial and temporal features from latent space during the upsampling. The extracted spatial and temporal features of each parallel node and adjacent node both are concatenated, i.e.,
\begin{equation}
S_{c}= Conc(S_{up},S_{d}), 
\end{equation}
where \textit{Conc} is concatenation operation. Then we perform separable convolution layers with $3 \times 3$ filters succeeded by the ReLU activation function to enhance the quality of the spatial-temporal feature vectors and reconstruct the original dimension of latent features vectors. Further, we apply the Sigmoid function using the convolution layer with $1 \times 1$ filters to simplify the spatial and temporal features, i.e.,
\begin{equation}
S_{m}= Sigmoid(S_{c}).
\end{equation}

The network has approximately 3.5 million trainable parameters. We notice that each layer of the VS-Net generates a feature map with a spatial structure in places of the video frames. Max Pooling layers use to increase the feature's map generation speed, which updates the weight matrix of the backbone models during the feature extraction from every separable convolution layer. During the bottom-up extraction of features from high to low resolution, upsampling operations with $2 \times 2$ filters are used to distribute the latent feature space and combine them with the previous layer and parallel layer's nodes' features. 

We used Approximation Rank Pooling (ARP) \cite{bilen2017action}, which takes input features from the intermediate layers of a VS-Net, trains on sub-sequences, and generates the output of a subspace. ARP not only gives low-rank approximation features, but it also conserves temporal order. The low-rank approximation differentiated and captured important characteristics of the data, which summarizes the document's position and orientation. Further, a quadratic ranking function captured the temporal order, which handles non-linear dependencies of the input features. Generally, the temporal order deals with the protuberances of the input channels onto the substance.

Due to the low-rank approximation, the down-sampling generates the generalized latent space vector. The sampling of mean and variance gives the efficient latent distribution of the VS-Net architecture. Based on the latent Gaussian distribution, the latent vector is generalized. The down-sampling and up-sampling are performed based on the variational encoder and decoder. For handling the over-fitting of the proposed model, the latent space of the down-sampling is normalized with the help of convolution layers and passed to up-sampling layers. After concatenating all the spatial and temporal features, we applied a convolution layer with $1 \times 1$ filters at the last node to generate feature vectors. At last, the previous node and parallel node features are aggregated and provide the saliency map.

\begin{table}[ht]
\centering
\caption{Performance comparison of the SOTA and proposed model (VS-Net) in terms of BCE+IoU loss($\%$), and testing speed (fps)}
\label{tab:1}
\resizebox{8.8cm}{!}{
\begin{tabular}{|c|c|c|c|c|}
\hline
\textbf{Input frame}& \textbf{Model} & \textbf{Output frame} &\textbf{Loss (BCE+IoU) }& \textbf{Testing speed (FPS)} \\
\hline
 & RNN+LSTM \cite{kompella2021semi}&\includegraphics[width=0.11\textwidth, height=9mm]{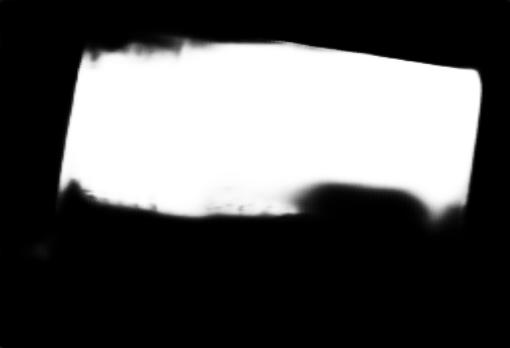}&0.032& 12.34\\
 \cline{2-5}
\multirow{9}*{\raisebox{-0.9\height}[0pt][0pt] { \includegraphics[width=0.11\textwidth, height=9mm]{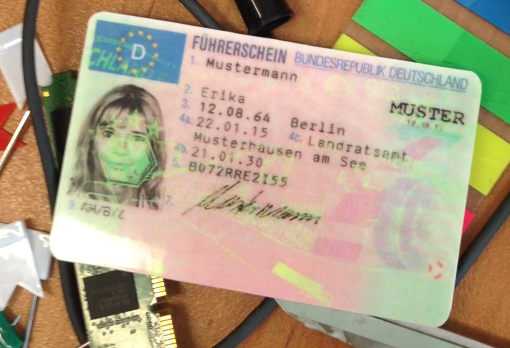}}}
 & U-Net \cite{liu2019u}&\includegraphics[width=0.11\textwidth, height=9mm]{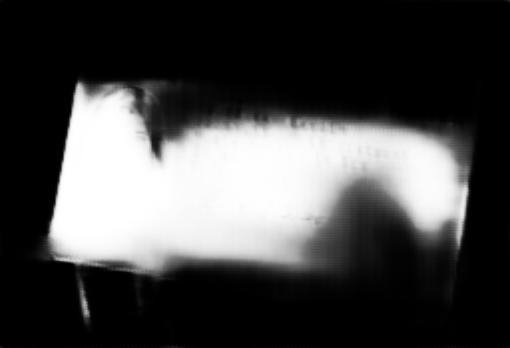}&0.028& 10.45\\
 \cline{2-5}
 & FCNN \cite{wang2017video}&\includegraphics[width=0.11\textwidth, height=9mm]{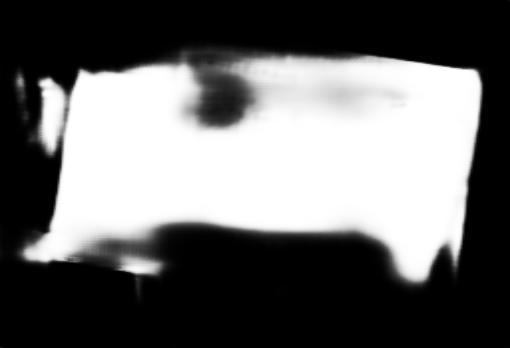}&0.038& 12.36\\
 \cline{2-5}
 & HE \cite{sheshku2020houghencoder}&\includegraphics[width=0.11\textwidth, height=9mm]{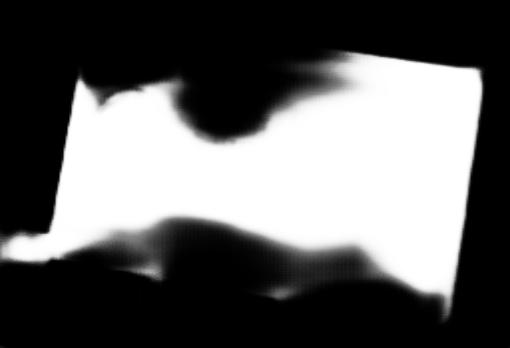}&0.041& 14.45\\
 \cline{2-5}
 & STCRF \cite{le2018video}&\includegraphics[width=0.11\textwidth, height=9mm]{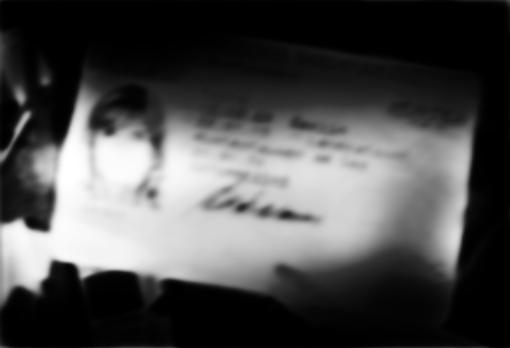}&0.039& 13.54\\
 \cline{2-5}
   & AED \cite{zhou2020salient}&\includegraphics[width=0.11\textwidth, height=9mm]{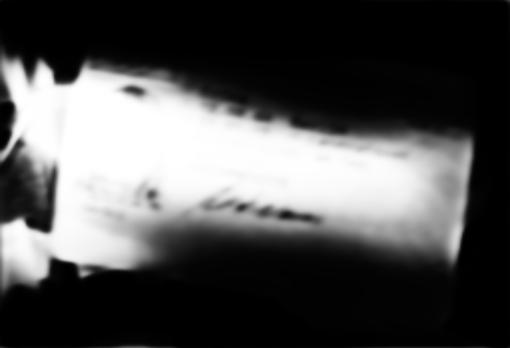}&0.035 & 12.89\\
 \cline{2-5}
& CNN+LSTM \cite{rhanoui2019cnn}&\includegraphics[width=0.11\textwidth, height=9mm]{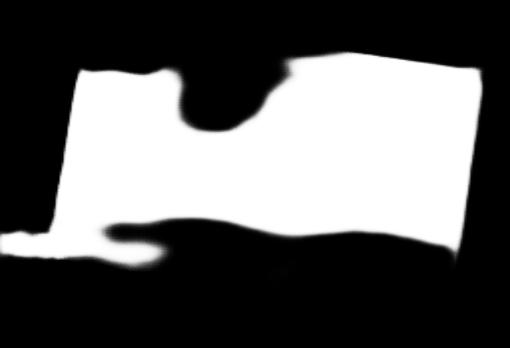}&0.036 & 12.63 \\
 \cline{2-5}
& RCNN \cite{sharma2019saliency}&\includegraphics[width=0.11\textwidth, height=9mm]{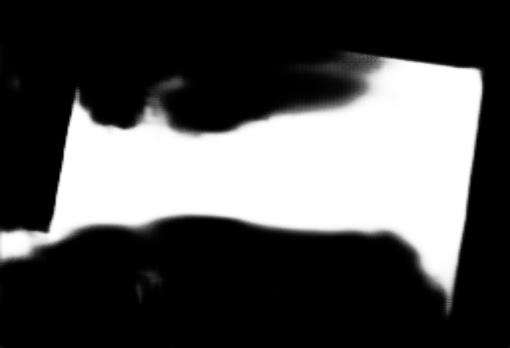}&0.025& 11.54\\
 \cline{2-5}
& \textbf{VS-Net}&\includegraphics[width=0.11\textwidth, height=9mm]{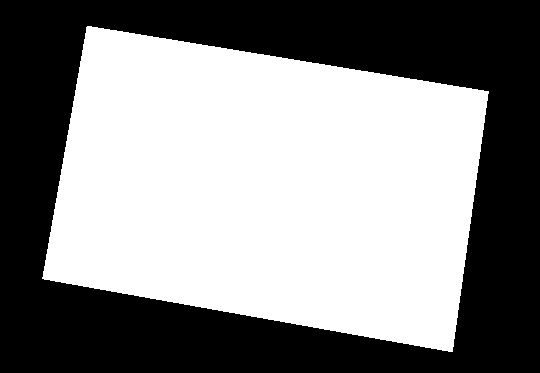}&\textbf{0.021}& \textbf{8.36}\\
\hline
\end{tabular}}
\end{table}

\subsection{Loss function}
During training, we use input frames $S_{k}$ with the corresponding ground-truth $G_{k}$ at frame t. The binary cross-entropy loss $L_{bce}$ \cite{ji2021full} is used to calculate the dissimilarity of the output and target, which is given as follows, 
\begin{equation} \label{eq:14}
\begin{split}
    l_{bce}(S_{k}, G_{k})= -\sum _{i,j}[G_{k}(i,j) log(S_{k}(i,j))+ \\ \newline (1-G_{k}(i,j))log(1-S_{k}(i,j))]
    \end{split}
\end{equation}

where (i,j) represents a coordinate of the frames. The IoU loss \cite{ahmed2015optimizing} is computed as: 
{\footnotesize
\begin{align}
 l_{IoU}(S_{k}, G_{k})=1-\frac{\sum_{i=1}^{w}\sum_{j=1}^{h}S_{k}(i,j)G_{k}(i,j)}{\sum_{i=1}^{w}\sum_{j=1}^{h}[S_{k}(i,j)+G_{k}(i,j)-S_{k}(i,j)G_{k}(i,j)]}
\end{align}\par} 
The total loss function is derived as,
\begin{equation}
\label{eq:16}
    Total_{loss}=l_{bce}(S_{k}, G_{k})+ \alpha l_{IoU}(S_{k}, G_{k}),
\end{equation}
where $\alpha$ is the weighting parameter for the IoU. For the final saliency map prediction, we utilize $S_{k}$ and $G_{k}$ since it shows that our experiments better utilized spatial-temporal cues.

\section{Experimental Setup and Result Analysis}
\subsection{MIDV-500 Dataset}
We have considered the MIDV-500 dataset \cite{arlazarov2019midv}, which contains video clips of 50 different identity documents (seventeen Id cards, fourteen passports, six identity documents, and thirteen driving licenses) of various countries. It has eight different variations of background and foreground scenes. The attributes of the dataset are TS (Table Scene), TA (Table Action), KS (Keyboard Scene), KA (Keyboard Action), HS (Hand Scene), HA (Hand Action), PS (Partial Scene), PA (Partial Action), CS (Complex Scene), CA (Complex Action). Thus a total of 500 videos are generated (50 documents$\times$ 5 desperation$\times$2 devices). Each video has a duration of three seconds, which is split into ten frames per second and the corresponding annotation. The dataset contains examples of multiple challenging scenarios, such as complex scenes, small objects with different variations of frames, appearances of background and foreground, cluttered background, etc. (a few examples are shown in \textit{Fig:}~\ref{figure:one}). The ground truth is prepared for each extracted video frame with various document locations in JSON format. It has 48 photo patches, 40 signature patches, and 546 text patches. The patches convert Cyrillic, Greek, Chinese, Japanese, Arabic, and Persian with singular Latin characters.

\subsection{Optimization and Propagation}
For preparing the VS-Net model, the Adaptive Moment Estimation (ADAM) optimizer is used to facilitate the computation of learning rates of each parameter using the first and second moment of the gradient. The saliency map optimization is cast as a “label propagation” problem, where uncertain labels are propagated based on background and foreground seeds. Further, we perform shift and rotation invariance robustly to the deformation of gray value variations. These random variations of deformation are sampled from the Gaussian distribution. Drop-out layers are used at the bottom stop and the previous level to normalize the latent space vectors. The proposed VS-Net model optimizes the loss of saliency maps from top-to-bottom and bottom-to-top and propagative seeds sequentially interact. The sequential seeding procedure optimizes feature map loss and improves the robustness to construct a saliency map. 

\subsection{Evaluation Metrics}
We used Intersection Over Union (IoU) loss as an evaluation metric, which is defined as the intersection over the predicted boundary box (bbox), and the actual bbox and divided with their union. A prediction considers True Positive if Intersection over union (IoU)$>$threshold, and False Positive If IoU$<$threshold \cite{ahmed2015optimizing}.
\begin{equation}
    IoU=\frac{ Area\quad of\quad Overlap }{Area\quad of\quad Union}\label{eq:9}
\end{equation}
The smaller the \textit{IoU} value, the better the performance. In our experiments, the threshold is set as 0.5. The comparison results in terms of accuracy are shown in \textit{Tab.} \ref{tab:2}).

\begin{table}[ht]
\centering
\caption{Comparison results with SOTA models and proposed model on MIDV-500 dataset}
\label{tab:2}
\begin{adjustbox}{max width=0.45\textwidth}
\begin{tabular}{l|l|l|l|l|l}
\hline
\multicolumn{6}{c}{Comparison of document detection accuracy ($\%$)} \\
\hline
Model & TS, & KS, & HS,& PS,& CS \\
     & TA&KA&HA&PA&CA\\
 \hline
U-Net (2019) \cite{liu2019u} & 96.44&97.43 &97.12 & 96.64 &97.40 \\
RNN+LSTM (2021) \cite{kompella2021semi} &97.12& 96.33 & 97.56 & 95.56 & 96.59\\
FCNN (2017)\cite{wang2017video} &97.78 & 98.39 & 98.89 & 96.43 & 97.43\\
HE  (2020) \cite{sheshku2020houghencoder}& 96.86 & 97.58 & 97.54& 96.32 & 96.45 \\
STCRF (2018) \cite{le2018video} & 97.43 & 96.98 & 97.56 & 96.43 & 97.98 \\
AED (2020)\cite{zhou2020salient}  & 96.97 & 98.54 & 98.74 & 96.78 & 97.69 \\
CNN+LSTM (2019)\cite{rhanoui2019cnn} & 97.56 & 98.54 & 98.74 & 96.78 & 97.69\\
RCNN (2019)\cite{sharma2019saliency} &98.43&98.69&97.78&97.19&98.56\\
\textbf{VS-Net}& \textbf{99.25}& \textbf{99.67} & \textbf{99.62} & \textbf{99.45} & \textbf{99.74}\\
\hline
\end{tabular}
\end{adjustbox}
\end{table}
\begin{table*}[ht]
\centering
\caption{Runtime comparison of SOTA models and VS-Net model}
\label{tab:3}
\begin{adjustbox}{max width=0.85\textwidth}
\begin{tabular}[h]{lccccccccc}
\toprule
&U-Net\cite{liu2019u}&RNN+LSTM\cite{kompella2021semi}&FCNN\cite{wang2017video}&HE \cite{sheshku2020houghencoder}&STCRF\cite{le2018video}&AED\cite{zhou2020salient}&CNN+LSTM\cite{rhanoui2019cnn}&RCNN\cite{sharma2019saliency}&\textbf{VS-Net}\\
\midrule
Runtime (s)&188 &191&190&198 & 200 & 199 & 196 & 189 &\textbf{175} \\
Step (ms)& 265 & 266 & 273 & 278 & 275 & 268 & 267 & 260 & \textbf{256}\\
\bottomrule
\end{tabular}
\end{adjustbox}
\end{table*}

\subsection{Implementation Details}
\subsubsection{Training Setup}
The experiments are accomplished on Intel Xeon(R) CPU E5-2640 v4 @ 2.40GHz X40 processor with 32GB RAM. We used Tensorflow-gpu2.0, Keras as the backend, and Python3.6 accelerated by NVIDIA RTX Graphic card (Quadro P5000 / PCIe /SSE2). Input frames are of size $256 \times 256$. The ADAM optimizer, weight decay of 0.006, batch size of 128, and learning rate of 0.001 are used to train the model, and BCE +IoU loss function is used to calculate the loss of the VS-Net model. The running time comparison is given in \textit{Tab.} \ref{tab:3}.

\subsubsection{Testing Setup and Runtime}
\begin{figure}[ht]
\centering
\includegraphics[width=8cm]{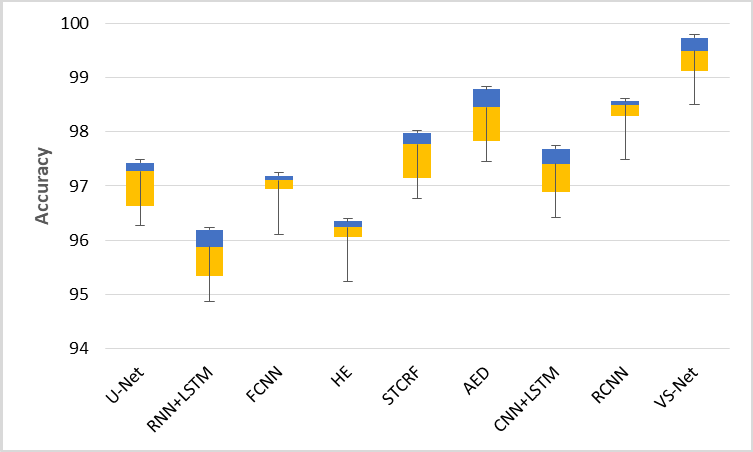}
\caption{5-fold cross-validation comparison result of VS-Net and SOTA models.}
\label{fig:three}
\end{figure}

We resize the frame $256 \times 256$  to feed into the corresponding branch for testing. The distribution of the dataset is in the ratio of 3:7. The average testing speed of our model is 9.46 fps which is less than the existing models. Additionally, we do not perform any pre-/post-processing \cite{sharma2019saliency}. For validating purpose, the 5-fold cross-validation are used to check the overfitting (see \textit{Fig.}\ref{fig:three}).

\begin{table}[ht]
\centering
\caption{Ablation Study with ResNet-50 as backbone.}
\label{tab:4}
\begin{adjustbox}{max width=0.5\textwidth}
\begin{tabular}[h]{c|c|c|c|c}
\hline
\textbf{$\#$ Param(M)} & \textbf{FLOPS(G)} & \textbf{Runtime (s)}&\textbf{Steps(ms)}& \textbf{Accuracy} \\
\hline
5.42& 10.2& 200 & 285 & 97.98 \\
3.54 & 7.3 & 185 & 275 & 98.85 \\
1.20 & 3.7 & 175 & 265 & 99.75 \\
0.95 & 2.5 & 165 & 255 & 96.45 \\
\hline
\end{tabular}
\end{adjustbox}
\end{table}

\textbf{Ablation Study} - The qualitative evaluation shows that our VS-Net model gives the best results on MIDV-500 \cite{arlazarov2019midv} datasets but is an extremely lightweight setting that has fewer parameters and FLOPs (see in \textit{Table.}~\ref{tab:4}).

\textbf{Speed Comparison}
The speed performance is calculated on a 64-bit Linux Ubuntu-18.04 operating system with Intel Core i7-4590 CPU @ 3.3 GHz. It has 32GB RAM and 1 TB Hard disk. The average speed is computed on all frame images of MIDV-500 datasets and does not include I/O file time. The parallel processing of multiple images is not allowed. Priorly, SOTA models require high computational powerful GPU system. For a fair comparison, the recent unsupervised models compare the speed on a normal CPU and the speed of deep learning methods reported on GPU. The speed comparison is given in \textit{Table} \ref{tab:3}. We show the trade-off between loss and testing speed in \textit{Table} \ref{tab:1}. Further, the average time cost of each step is examined. VS-Net consumes 256 ms and SOTA models 270 ms, respectively, while saliency prediction and global optimization use 8.36 ms and 10.0 ms.

\section{Conclusion and Future Work}
In this paper, we proposed a novel efficient and fast VS-Net model that fully leverages the spatiotemporal features to detect the video identity document. The proposed model used separable convolutions with variational autoencoder architecture, which performs downsampling for extracting the features and upsampling operation for decoding the latent space features. The Approximation Rank Pooling (ARP) is used low-rank approximation on frames to preserve the temporal locality. The bottom-end the generalized latent space is generated. Further, These features are combined to generate in fuse spatial and temporal characteristics. We validated each module of the proposed model with the help of extensive experiments, which can be considered as the unified solution advancing VSDD.

\bibliographystyle{IEEEtran}
\bibliography{IEEEfull}

\end{document}